\documentclass[10pt,twocolumn,letterpaper]{article}

\usepackage{cvpr}
\usepackage{times}
\usepackage{epsfig}
\usepackage{graphicx}
\usepackage{amsmath}
\usepackage{amssymb}

\usepackage{amsfonts}
\usepackage{graphicx}
\usepackage{subfig}
\usepackage{multirow}
\usepackage{booktabs}
\usepackage{algorithm}
\usepackage{algorithmic}
\usepackage{CJK}
\usepackage{color}
\usepackage{amsthm}
\usepackage{setspace}
\usepackage{bm}
\usepackage{caption}
\usepackage{verbatim}
\usepackage{enumerate}
\usepackage{cite}

\usepackage{float}
\makeatletter
\newcommand*{\rom}[1]{\expandafter\@slowromancap\romannumeral #1@}

\usepackage[pagebackref=true,breaklinks=true,letterpaper=true,colorlinks,bookmarks=false]{hyperref}

 \cvprfinalcopy 

\def\cvprPaperID{1995} 

\begin{document}
\pagenumbering{gobble}
\title{Multi-granularity Generator for Temporal Action Proposal}
\date{}
\author{Yuan Liu$^\sharp$\thanks{This work was done while Yuan Liu was a Research Intern with Tencent AI Lab. } \qquad
Lin Ma$^\natural$\thanks{Corresponding authors.}\qquad
Yifeng Zhang$^{\sharp\dagger}$ \qquad
Wei Liu$^\natural$ \qquad
Shih-Fu Chang$^\S$ \\
$^\natural$Tencent AI Lab \qquad $^\sharp$Southeast University \qquad $^\S$Columbia University\\ 
{\tt\small \{lhy19930911,forest.linma\}@gmail.com\qquad yfz@seu.edu.cn\qquad  \{wl2223,sc250\}@columbia.edu}
}
\maketitle

\begin{abstract}
Temporal action proposal generation is an important task, aiming to localize the video segments containing human actions in an untrimmed video.
In this paper, we propose a multi-granularity generator (MGG) to perform the temporal action proposal from different granularity perspectives, relying on the video visual features equipped with the position embedding information. First, we propose to use a bilinear matching model to exploit the rich local information within the video sequence. Afterwards, two components, namely  segment proposal producer (SPP) and  frame actionness producer (FAP), are combined to perform the task of temporal action proposal at two distinct granularities. SPP considers the whole video in the form of feature pyramid and generates segment proposals from one coarse perspective, while FAP carries out a finer actionness evaluation for each video frame. Our proposed MGG can be trained in an end-to-end fashion. By temporally adjusting the segment proposals with fine-grained frame actionness information, MGG achieves  the superior performance over state-of-the-art methods on the public THUMOS-14 and
ActivityNet-1.3 datasets. Moreover, we employ existing action classifiers to perform the classification of the proposals generated by MGG, leading to significant improvements compared against the competing methods for the video detection task.
\end{abstract}

\begin{figure}[htbp]
\centering
\includegraphics[width=8cm]{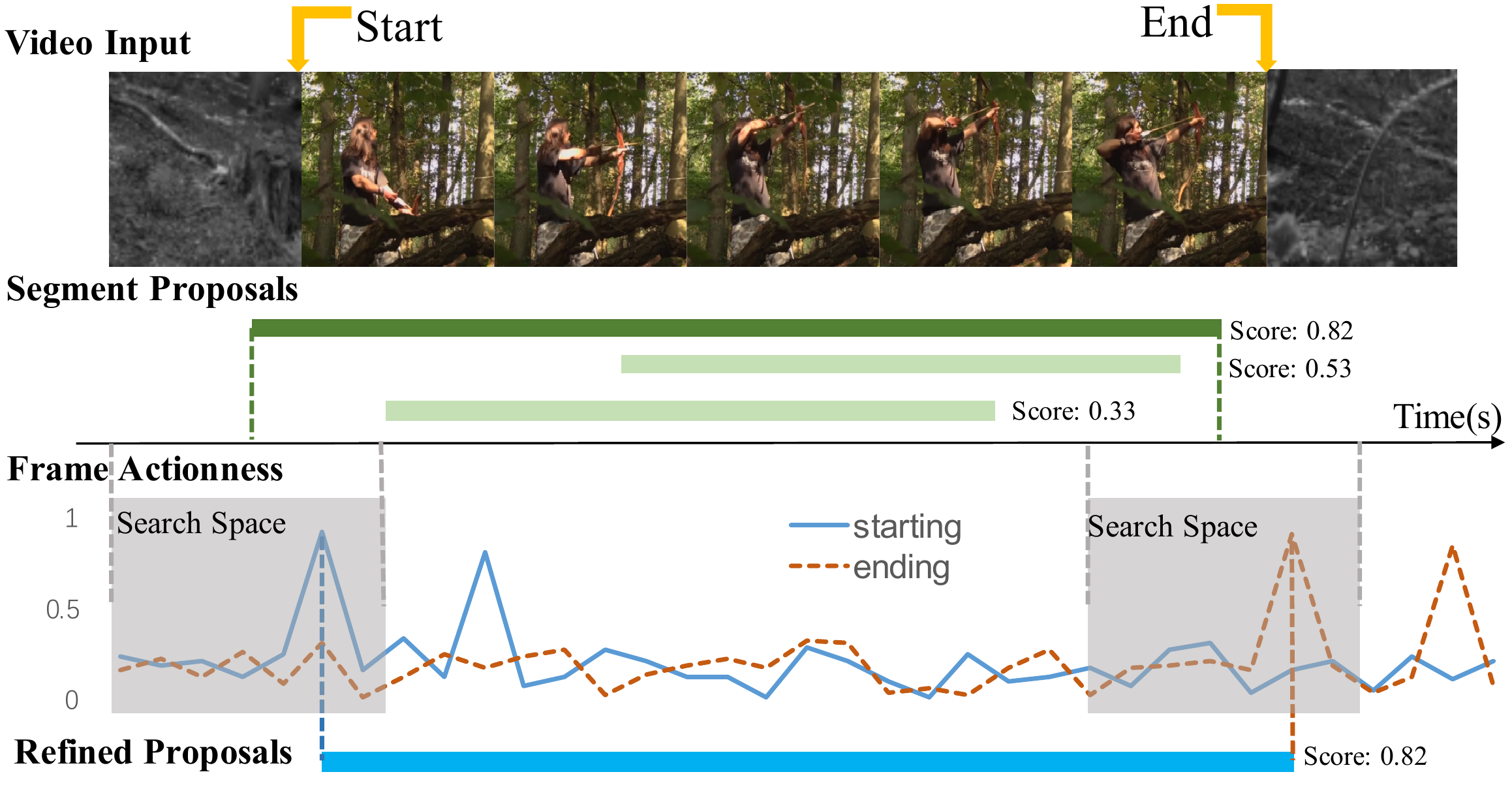}
\caption{
Our proposed MGG can generate segment proposals and frame actionness simultaneously, which  helps discover information about possible actions at both the coarse and fine levels. By temporally adjusting the boundaries of the segment within the search space determined by the computed frame actionness, MGG can  yield refined action proposals with both high recall and precision.
 } 
 \label{boundary}
\end{figure} 

\section{Introduction}
Temporal action proposal \cite{turn_tap,dap} aims at capturing video temporal intervals that are likely to contain an action in an untrimmed video. This task plays an important role in video analysis and can thus be applied in many areas, such as 
action recognition~\cite{wenbing,jiang2015human,jiang2012trajectory-based,cao2013mining}, 
summarization~\cite{highlight,summar}, grounding~\cite{chen2018temporally,chen2019localizing} and  captioning~\cite{wang2018bidirectional,wang2018reconstruction}. Many methods~\cite{TAG,ctap} have been proposed to handle this task, and have shown that, akin to object proposals for object detection~\cite{faster_rcnn}, temporal action proposal has a crucial impact on the quality of action detection. 

High-quality action proposal methods should capture temporal action instances with both high  recall and high temporal overlapping with ground-truths, meanwhile producing proposals without many false alarms.
One type of existing methods focuses on generating segment proposals~\cite{shou_action,turn_tap}, where the initial segments are regularly distributed or manually defined over the video sequence. A binary classier is thereafter trained to evaluate the confidence scores of the segments. Such methods are able to generate proposals of various temporal spans. However, since the segments are regularly distributed or manually defined, the generated proposals naturally have  imprecise boundary information, even though boundary regressors are further applied. Another thread of work, like \cite{TAG,cdc,sms}, tackles the action proposal task in the form of evaluating frame actionness. These methods densely evaluate the confidence score for each frame and group consecutive frames together as candidate proposals. 
The whole video sequence is analyzed at a finer level, in contrast with the segment proposal based methods. As a result,  the boundaries of the generated proposals are of high precision. However, such methods often produce low confidence scores for long video segments, resulting in misses of true action segments and thus low recalls.

Obviously, these two types of methods are complementary to each other. Boundary sensitive network (BSN)~\cite{bsn} adopts a ``local to global" scheme for action proposal, which locally detects the boudary information and globally ranks the candidate proposals. Complementary temporal action proposal (CTAP)~\cite{ctap}   consists of three stages, which are initial proposal generation, complementary
proposal collection, and boundary adjustment and proposal ranking, respectively.
However,
both of these two methods are multi-stage models with the modules in
different stages trained independently, without overall
optimization of  the models. Another drawback is the neglect of the temporal position information, which conveys the temporal ordering information of the video sequence and is thereby expected to be helpful for precisely localizing the proposal boundary.

In order to address the aforementioned drawbacks, we propose a multi-granularity generator (MGG) by taking full advantage of both segment proposal and frame actionness based methods. At the beginning, the frame position embedding, realized with cosine and sine functions of different wavelengths, is combined with the video frame features.  The combined features are then fed to  MGG to perform the temporal action proposal. Specifically,  a bilinear matching model is first proposed to exploit the rich local information of the video sequence. Afterwards, two components, namely segment proposal producer (SPP) and frame actionness producer (FAP), are coupling together and responsible for generating coarse segment proposals and evaluating fine frame actionness, respectively. SPP uses a U-shape architecture with lateral connections to generate candidate proposals of different temporal spans with high recall. For FAP, we densely evaluate  the probabilities of each frame being the starting point, ending point, and inside a correct proposal (middle point). During the inference, MGG can further temporally adjust the segment boundaries with respect to the frame actionness information as shown in Fig.~\ref{boundary}, and consequently produce refined action proposals.

In summary, the main contributions of our work are four-fold:
\begin{itemize}
    \item We propose an end-to-end multi-granularity generator (MGG) for temporal action proposal, using a novel representation integrating video features and the position embedding information. MGG simultaneously generates coarse segment proposals by perceiving the whole video sequence, and predicts the frame actionness by densely evaluating each video frame. 
    \item A bilinear matching model is proposed to exploit the rich local information within the video sequence, which is thereafter harnessed by the following SPP and FAP.
    \item SPP is realized in a U-shape architecture with lateral connections, capturing temporal proposals of various spans with high recall, while FAP evaluates the probabilities of each frame being the stating point, ending point, and middle point.
    \item Through temporally adjusting the segment proposal boundaries using the complementary information in the frame actionness, our proposed MGG achieves the state-of-the-art performances on the THUMOS-14 and ActivityNet-1.3 datasets for the temporal action proposal task. 
\end{itemize}

\section{Related Work}
A large number of existing approaches have been proposed to tackle the problem of temporal action detection \cite{buch2017end,ssn,yeung2016end,ssad,autoloc,lang_model}. Inspired by the success of two-stage detectors like RCNN \cite{rcnn}, many recent methods adopt a proposal-plus-classification framework \cite{r-c3d,re_faster,cdc,tcn}, where classifiers are applied on a smaller number of class agnostic
segment proposals for detection. The proposal stage and classification stage can be trained separately \cite{cdc,shou_action,ssn} or jointly \cite{re_faster,r-c3d}, and demonstrate very competitive results. Regarding temporal action proposal, DAP \cite{dap} and SST
\cite{sst} introduce RNNs to process video sequences in a single
pass. However, LSTM \cite{lstm} and GRU \cite{gru} fail to handle video segments with long
time spans.
Alternatively, \cite{slide_window,tcn,shou_action} directly generate proposals from sliding windows.
R-C3D \cite{r-c3d} and TAL-Net \cite{re_faster} follow the Faster R-CNN \cite{faster_rcnn} paradigm to
predict locations of temporal proposals and the corresponding categories. These methods perceive the whole videos in a coarser level, while the pre-defined temporal intervals may limit the accuracy of generated proposals.
Methods like temporal action grouping (TAG) \cite{TAG} and CDC \cite{cdc} produce final proposals by densely giving evaluation to each frame. Analyzing videos in a finer level,  the generated proposals are quite accurate in boundaries.
In our work, MGG tackles the problem of temporal action proposal in both coarse and fine perspectives, being better at both recall and overlapping.

\begin{figure}
\center
\includegraphics[width=8.3cm]{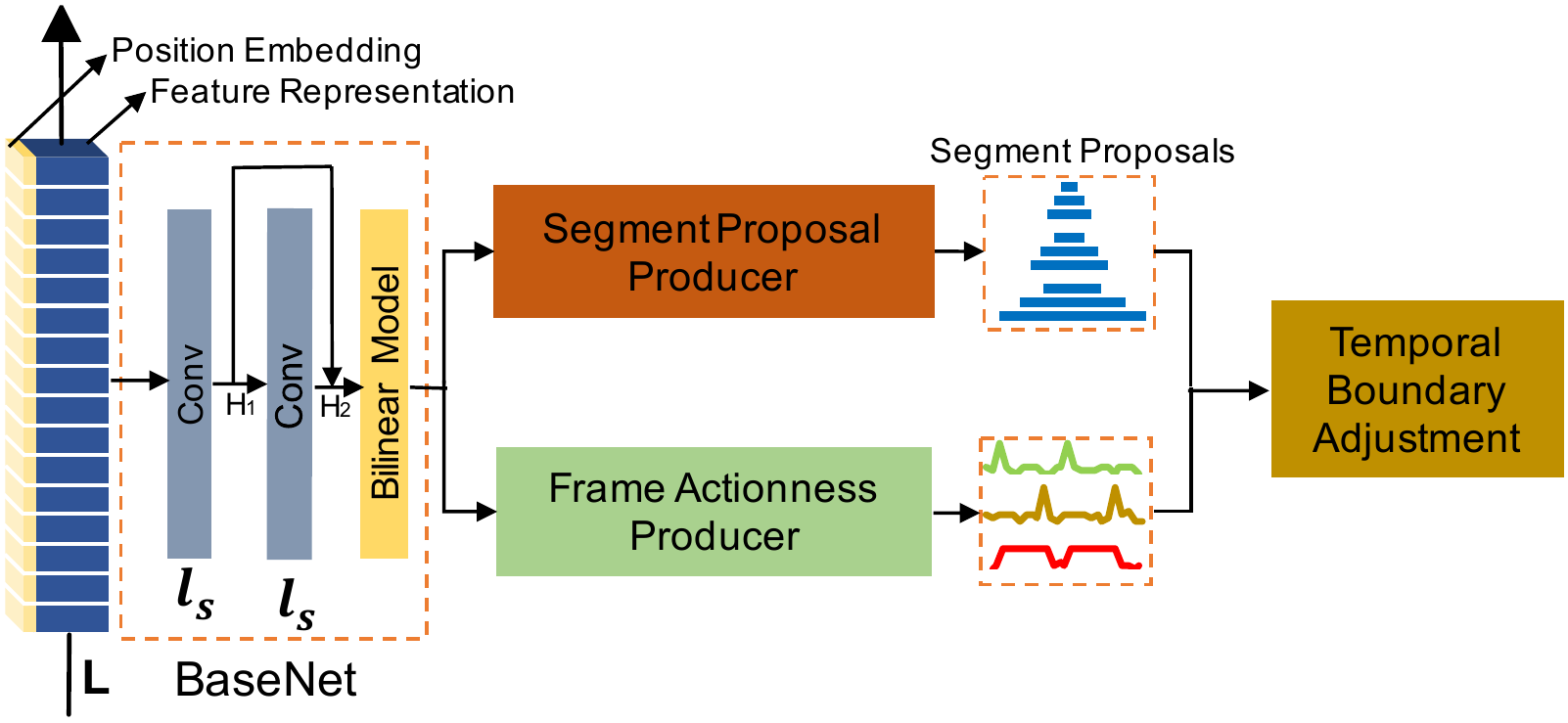}
\caption{The architecture of our proposed MGG. The video visual features are first combined with the position embedding information to form the video representations. The proposed BaseNet relies on a blinear model to exploit the rich local information within the sequential video representations. Segment proposal producer ({SPP}) is realized by using a U-shape architecture with lateral connections to generate proposals of different temporal lengths, while frame actionness producer ({FAP}) evaluates each frame whether it is the starting point, ending point, or middle point. 
With the temporal boundary adjustment (TBA) module, boundaries of the segment proposals are temporally adjusted based on computed frame actionness, and the refined accurate action proposals are therefore generated.} 
\label{network}
\end{figure}

\section{Our Approach}
\label{sec:approach}

Given an untrimmed video sequence $\textbf{s}=\{s_n\}^{l_s}_{n=1}$ with its length as $l_s$, temporal action proposal aims at detecting action instances $\varphi_{p}=\{\xi_{n}=[t_{s,n},t_{e,n}] \}_{n=1}^{M_{s}} $, where 
$M_{s}$ is the total number of action instances, and  $[t_{s,n},t_{e,n}]$ denote the starting and ending points of an action instance $\xi_{n}$, respectively.    

We propose one novel neural network, namely MGG shown in Fig.~\ref{network}, which analyzes the video and performs temporal action proposal at different granularities. 
Specifically, our proposed MGG consists of four components. The video visual features are first combined with the position embedding information to yield the video representations.  The subsequent BaseNet relies on a blinear model to exploit the rich local information within the sequential video representations. Afterwards, SPP and FAP are used to produce the action proposals from the coarse (segment) and fine (frame) perspectives, respectively. Finally, the temporal boundary adjustment (TBA) module adjusts the segment proposal boundaries regarding the frame actionness and therefore generates action proposals of both high recall and precision.

\subsection{Video Representation} 
First, we need to encode the video sequence and generate the corresponding representations. Same as the previous work~\cite{bsn,ctap}, one convolutional neural network (CNN) is used to convert one video sequence $\textbf{s}=\{s_n\}^{l_s}_{n=1}$ into one visual feature sequence $\textbf{f}=\{f_n\}^{l_s}_{n=1}$ with $f_n \in R^{d_f}$. $d_f$ is the dimension of each feature representation. However, the temporal ordering information of the video sequence is not considered. Inspired by~\cite{attention_is,conv_seq2seq}, we embed the position information to explicitly characterize the ordering information of each visual feature, which is believed to benefit the action proposal generation. {The position information of the $n$-$th$ $(n\in[1,l_{s}])$ visual feature $f_n$ is embedded into a feature $p_{n}$ with a dimension $d_p$ by computing cosine and sine functions of different wavelengths:
\begin{equation}
\begin{split}
    p_n(2i) &= \text{sin}(n/10000^{2i/d_p}),\\
    p_n(2i+1) &= \text{cos}(n/10000^{2i/d_p}),
\end{split}
\end{equation}
where $i$ is the index of the dimension. The generated position embedding $p_n$ will be equipped with the visual feature representation $f_{n}$ via concatenation, denoted by $l_{n} = [f_{n},p_{n}]$. As such, the final video representations $L = \{l_{n}\}_{n=1}^{l_{s}} \in R^{l_{s}\times d_{l}}$ are obtained, where $d_{l} = d_{f}+d_p$ denotes the dimension of the fused representations.}

\subsection{BaseNet}
Based on the video representations, we propose a novel BaseNet to exploit the rich local behaviors within the video sequence. As shown in Fig.~\ref{network}, two temporal convolutional layers are first stacked to exploit video temporal relationships. A typical temporal convolutional layer is denoted as $\text{Conv}(n_f,n_k,\Omega )$, where $n_f$, $n_k$, and $\Omega $ are filter numbers, kernel size, and activation function, respectively. In our proposed BaseNet, the two convolutional layers are of the same architecture, specifically $\text{Conv}(d_h,k,\text{ReLU})$, where $d_h$ is set to 512, $k$ is set to 5, and ReLU refers to the activation of rectified linear units~\cite{relu}. The outputs of these two temporal convolutional layers are denoted as  $H_1$ and $H_2$, respectively. 

The intermediate representations $H_{1}$ and $H_{2}$ express the semantic information of the video sequence at different levels, which are rich in characterizing the local information. We propose a bilinear matching model~\cite{bilinear_old} to capture the interaction behaviors between $H_1$ and $H_2$. Due to a large number of parameters contained in a traditional bilinear matching model, which result in an increased computational complexity and a higher convergence difficulty, 
we turn to pursue a  factorized bilinear matching model~\cite{bilinear_new,fengyang}:
\begin{equation}
\begin{split}
    \hat{H}_{1}^{n} =& H_{1}^{n}W_{i}+b_{i},\\
    \hat{H}_{2}^{n} =& H_{2}^{n}W_{i}+b_{i},\\
    T_{i}^{n} =& \hat{H}_{1}^{n}\hat{H}_{2}^{n\top},
\end{split}
\end{equation}
where $H_{1}^{n}\in R^{1\times d_{h}}$ and $H_{2}^{n}\in
R^{1\times d_{h}}$ denote the corresponding representations at the $n$-$th$ location of $H_{1}$ and $H_{2}$, respectively. $W_{i}\in R^{d_{h}\times g}$ and $b_{i}\in R^{1\times g}$ are the parameters to be learned, with $g$ denoting a hyperparameter and being much smaller than $d_{h}$. Due to the smaller value of $g$, fewer parameters are introduced, which are easier for training. As such, the matching video representations $T=[T^{1},..,T^{l_s}]$, with $T^{n}=[T_{1}^{n},T_{2}^{n},..,T_{d_{h}}^{n}]$ denoting the $n$-$th$ feature, is obtained and used as the input to the following SPP and FAP for proposal generation.

\begin{figure}[!t]
\centering
\subfloat[The architecture of SPP]{\includegraphics[scale=0.52]{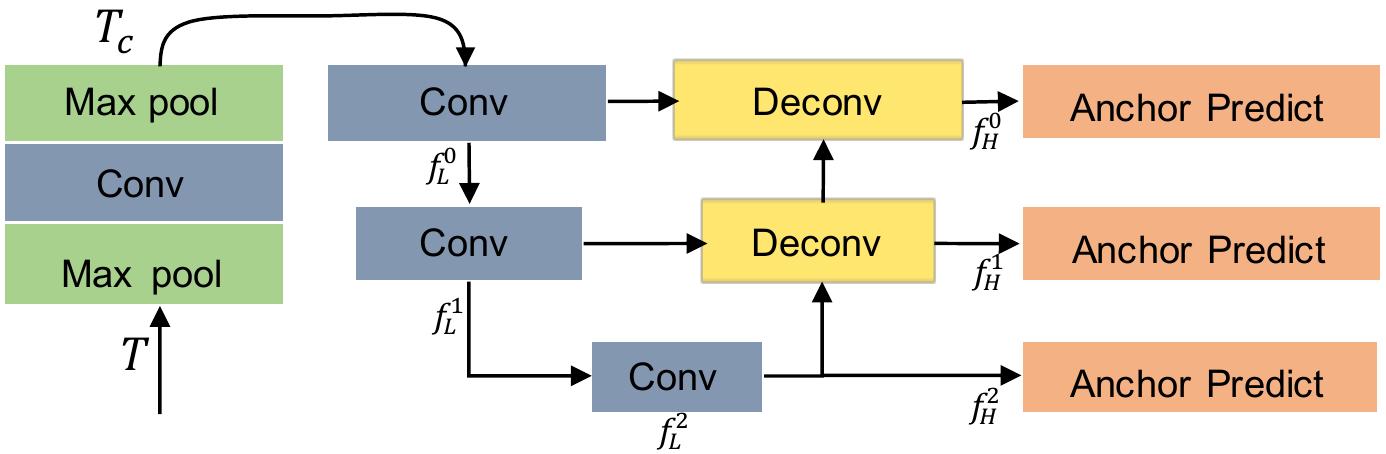}}\\
\subfloat[The anchor predict module]{\includegraphics[scale=0.49]{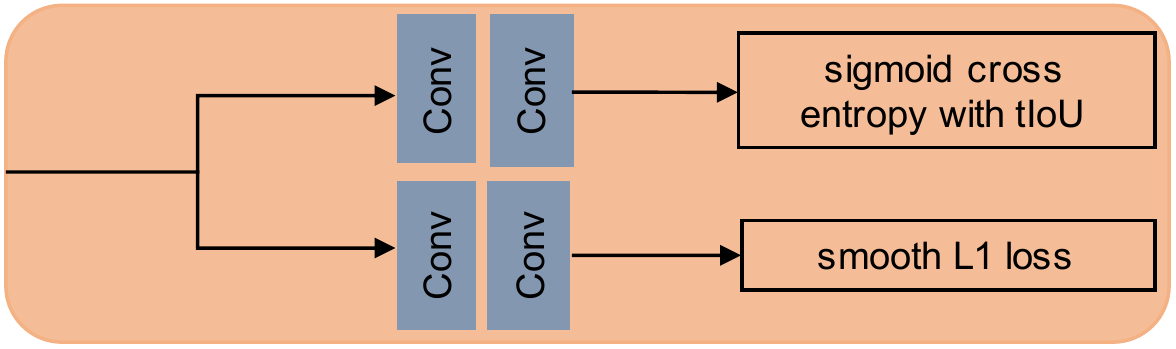}}
\caption{(a) Overview of SPP with pyramid levels $M=3$. With a U-shape architecture and lateral connections, the generated  feature pyramid $F_H$ is helpful for capturing proposals with different temporal durations. (b) The anchor predict module has two branches which are used for classification and boundary regression, respectively.}
\label{segment actionness}
\end{figure}

\subsection{Segment Proposal Producer}
\label{section3.3}

Due to large variations of action duration, capturing proposals of different temporal lengths with high recall is a big challenge. Xu \textit{et al.}~\cite{r-c3d} used one feature map to locate proposals of various temporal spans, yielding low average recall. SSAD~\cite{ssad} and TAL-Net~\cite{re_faster} use a feature pyramid network, with each layer being responsible for proposal localization with specific time spans. However, each pyramid layer, especially the lower ones being unaware of high-level semantic information, is unable to localize  temporal proposals accurately. To deal with this issue, we adopt a U-shape architecture with lateral connections between the convolutional and deconvolutional layers, as shown in Fig.~\ref{segment actionness}.

With yielded matching video representations $T$ as input, SPP first stacks three layers, specifically one temporal convolutional layer and two max-pooling layers, to reduce the temporal dimension and hence increase the size of the receptive field accordingly. As a result, the temporal feature $T_c$ with temporal dimension $l_s/8$ is taken as the input of the U-shape architecture.

Same as the previous work, such as Unet~\cite{unet}, FPN~\cite{fpn}, and DSSD \cite{dssd}, our U-shape architecture also consists of a contracting path and an expansive path as well as the lateral connections. Regarding the contracting path, with repeated temporal convolutions with stride 2 for downsampling, the feature pyramid (FP) $F_L= \{f_L^{(0)},f_L^{(1)},...f_L^{(M-1)}\}$ is obtained, where $f_L^{(n)}$ is the $n$-$th$ level  feature map of $F_L$ with  temporal dimension $\frac{l_s}{8*2^n}$. $M$ denotes the total number of pyramid levels.
For the expansive path, temporal deconvolutions are adopted on multiple layers with an upscaling factor of 2. Via lateral connections, high-level
features from the expansive path are combined with the corresponding low-level features, with the fused features denoted as $f_H^{(n)}$. Repeating this operation, the fused feature pyramid is defined as $F_H=\{f_H^{(0)},f_H^{(1)},...f_H^{(M-1)}\}$.  Different levels of feature pyramids are of different receptive fields, which are responsible for locating proposals of different temporal spans. 

 A set of anchors are regularly distributed over each level of feature pyramid $F_H$, based on which segment proposals are produced.  As shown in Fig.~\ref{segment actionness}, each $f_H$ is followed by two branches, with each branch realized by stacking two layers of temporal convolutions. Specifically, one branch is the classification module to predict the probability of a ground-truth proposal being present at each temporal location for each of the $\rho$ anchors, where $\rho$ is the number of anchors per location of the feature pyramid. The other branch is the boundary regression module to yield the relative offset between the anchor and the ground-truth proposal.

\subsection{Frame Actionness Producer}
Based on the yielded matching video representations $T$, the frame actionness producer (FAP) is proposed to evaluate the actionness  of each frame. Specifically, three two-layer temporal convolutional networks are used to generate the starting point, ending point, and middle point probabilities for each frame, respectively. Please note that two-layer temporal convolutional networks share the same configuration, where the first one is defined as $\text{Conv}(d_f,k,\text{ReLU})$ and the second one is  $\text{Conv}(1,k,\text{Sigmoid})$. $d_f$ is set to 64, while $k$, as the kernel size, is set to 3. And their weights are not shared. As a result, we obtain three probability sequences, namely the  starting probability sequence $P_{s}=\left\{p_{n}^{s}\right\}_{n=1}^{l_{s}}$, the ending probability sequence
$P_{e}=\left\{p_{n}^{e}\right\}_{n=1}^{l_{s}}$, and
the middle probability sequence $P_{m}=\left\{p_{n}^{m}\right\}_{n=1}^{l_{s}}$, with  $p_{n}^{s}$,  $p_{n}^{e}$, and $p_{n}^{m}$ denoting the starting,  ending, and middle probabilities of the $n$-$th$ feature, respectively. Compared with the generated segment proposals by SPP, the frame actionness yielded by FAP densely evaluates each frame in a finer manner.

\section{Training and Inference}

In this section, we will first introduce how to train our proposed MGG network, which can subsequently generate segment proposals and frame  actionness. During the inference, we propose one novel fusion strategy by temporally adjusting the  segment boundary information with respect to the frame actionness. 
 
\subsection{Training}
As introduced in Sec.~\ref{sec:approach}, our proposed MGG considers both the SPP and FAP together with a shared BaseNet. During the training process, these three components cooperate with each other and are jointly trained in an end-to-end fashion. Specifically, the objective function of our proposed MGG is defined as:
\begin{equation}
L_{MGG} = L_{SPP}+\beta L_{FAP},
\end{equation}
where $L_{SPP}$ and $L_{FAP}$ are the objective functions defined for SPP and FAP, respectively. $\beta$ is a parameter to adjust their relative contributions, which is empirically set to $0.1$. Detailed information about $L_{SPP}$ and $L_{FAP}$ will be introduced in what follows.

\subsubsection{SPP Training}
 Our proposed SPP produces a set of anchor segments  for each level  of the fused feature pyramids $F_H$. We  first introduce how to assign labels to the corresponding anchor segments. Subsequently, the objective function by referring to the assigned labels is introduced.

\textbf{Label Assignment.} 
{Same as Faster RCNN~\cite{faster_rcnn}, we assign a binary class label to each anchor segment. A positive label is assigned if it overlaps with some ground-truth proposals with temporal Intersection-over-Union (tIoU) higher than 0.7, or has the highest tIoU with a ground-truth proposal. Anchors are regarded as negative if the maximum tIoU with all ground-truth proposals is lower than 0.3. Anchors that are neither positive nor negative are filtered out. To ease the issue of class imbalance, we sample the positive and negative examples with a ratio of 1:1 for training.}

\textbf{Objective Function.} As shown in  Fig.~\ref{segment actionness} (b), we perform a multi-task training for SPP, which not only predicts the actionness of each anchor segment but also regresses its boundary information. For actionness prediction, the cross-entropy function is used, while the smooth $L_1$ loss function, as introduced in~\cite{fast_rcnn}, is used for boundary regression. Specifically, the objective function is defined as:
\begin{equation}
\begin{split}
L_{SPP} = & \frac{1}{N_{cls}}\sum_{i}L_{cls}(p_{i},p_{i}^{*})+\\
&\gamma \frac{1}{N_{reg}}\sum
_{i}[p_{i}^{*}\geqslant 1]L_{reg}(W_{i},W_{i}^{*}),
\end{split}
\end{equation}
where $\gamma$ is a trade-off parameter, which is set to $0.001$ empirically. $N_{cls}$ is the total number of training examples. $p_{i}$ stands for the yielded score. $p_i^*$ is the label, 1 for positive samples and 0 for negative samples. $L_{cls}$ is the cross-entropy loss function between $p_{i}$ and $p_{i}^{*}$. The smooth $L_1$ loss function $L_{reg}$ is activated only when the ground-truth label $p_{i}^{*}$ is positive, and disabled otherwise. $N_{reg}$ is the number of training examples whose $p_{i}^{*}$ is positive. $W_{i}=\left \{t_{c},t_{l}\right\}$ represents the predicted relative offsets of anchor segments. $W_{i}^{*}=\left\{t_{c}^{*},t_{l}^{*}\right\}$ indicates the relative offsets between ground-truth proposals and the anchors, which can be computed: 
\begin{equation}
\left\{\begin{matrix} t_{c}^{*}=&(c_{i}^{*}-c_{i})/ l_{i},  \\
t_{l}^{*}=&log(l_{i}^{*}/l^{i}), \end{matrix}\right.
\end{equation}
where $c_{i}$ and $l_{i}$ indicate the center and  length of anchor segments, respectively. $c_{i}^{*}$ and $l_{i}^{*}$ represent the center and length  of the ground-truth action instances.

\subsubsection{ FAP Training}
FAP takes the matching video representations with their length as $l_{s}$  as input and outputs three probability sequences, namely the starting probability sequence 
$P_{s}=\left\{p_{n}^{s}\right\}_{n=1}^{l_s}$,  the ending probability sequence
$P_{e}=\left\{p_{n}^{e}\right\}_{n=1}^{l_s}$, and the middle probability sequence
 $P_{m}=\left\{p_{n}^{m}\right\}_{n=1}^{l_s}$.

\textbf{Label Assignment.} The ground-truth annotations of temporal action proposals are denoted as $\pi =
\left\{\psi_{n}=[t_{s,n},t_{e,n}]\right\}_{n=1}^{M_{a}}$, where
$M_{a}$ is the total number of annotations. For each action instance
$\psi_{n}\in\pi$, we define the starting, ending, and middle
regions as $[t_{s,n}-d_{d,n}/\eta,t_{s,n}+d_{d,n}/\eta], [t_{e,n}-d_{d,n}/\eta,t_{e,n}+d_{d,n}/\eta]$, and $[t_{s,n},t_{e,n}]$,
respectively, where $d_{d,n}=t_{e,n}-t_{s,n}$ is the duration of the
annotated action instance and $\eta$ is set to 10 empirically. For each visual feature, if it lies in the starting, ending, or
middle regions of any action instances, its corresponding
starting, ending, or middle label will be set to 1, otherwise 0. In this
way, we obtain the ground-truth label for the three sequences, which are denoted
as
 $G_{s}=\left\{g_{n}^{s}\right\}_{n=1}^{l_s}$, $G_{e}=\left\{g_{n}^{e}\right\}_{n=1}^{l_s}$, and $G_{m}=\left\{g_{n}^{m}\right\}_{n=1}^{l_s}$, respectively.

 \textbf{Objective Function.} Given the predicted probability sequences and ground-truth labels, the objective function for FAP is defined as:
 \begin{equation}
L_{FAP}^{all} = \lambda_{s}L_{FAP}^{s}+\lambda_{e}L_{FAP}^{e}+\lambda_{m}L_{FAP}^{m}.
\end{equation}
The cross-entropy loss function is used for calculating  all the  three losses $L_{FAP}^{s}$, $L_{FAP}^{e}$, and $L_{FAP}^{m}$, where a weighting factor set by an inverse class frequency is introduced to address class imbalance. $L_{FAP}^{all}$ is the sum of the starting loss $L_{FAP}^{s}$, ending loss $L_{FAP}^{e}$,
and middle loss $L_{FAP}^{m}$, where $\lambda_{s}$, $\lambda_{e}$, and
$\lambda_{m}$ are the weights specifying the relative importance
of each part. In our experiments, we set $\lambda_{s} =
\lambda_{e}=\lambda_{m}=1$.

\subsection{Inference} 
\label{TBA}
As aforementioned, SPP aims to locate segment proposals of various temporal spans, thus yielding segment proposals with inaccurate boundary information. On the contrary, FAP gives an evaluation of each video frame in a finer level, which makes it sensitive to boundaries of action proposals.
Obviously, SPP and FAP are complementary to each other. Therefore, during the inference phase, we propose the temporal boundary adjustment (TBA) module realized in a  two-stage fusion strategy to improve the boundary accuracy of segment proposals with respect to the frame actionness.

\textbf{Stage \rom{1}.} 
We first use non-maximum suppression (NMS) to post-process the segment-level action instances detected by SPP. The generated results are denoted as $\varphi_{p}=\{\xi_{n}=[t_{s,n},t_{e,n}] \}_{n=1}^{M_{s}}$, where $M_{s}$ is the total number of the detected action instances, and $t_{s,n}$ and $t_{e,n}$ denote the corresponding starting and ending times of an action instance $\xi_{n}$, respectively. We will adjust $t_{s,n}$ and $t_{e,n}$ by referring to the starting and ending scores detected in FAP. Firstly, we set two context regions $\xi_{n}^{s}$ and $\xi_{n}^{e}$, which are named as the searching space:
\begin{equation}
\begin{split}
\xi_{n}^{s}= [t_{s,n}-d_{d,n}/\varepsilon ,t_{s,n}+d_{d,n}/\varepsilon ],\\
\xi_{n}^{e}= [t_{e,n}-d_{d,n}/\varepsilon ,t_{e,n}+d_{d,n}/\varepsilon ],
\end{split}
\end{equation}
where $d_{d,n}=t_{e,n}-t_{s,n}$ is the duration of $\xi_{n}$. $\varepsilon$ which controls the size of the searching space is set to 5 . The max starting score and the corresponding time in the region of $\xi_{n}^{s}$ are defined as $c_{n}^{s}$ and $t_{s,n}^{max}$, respectively , and the max ending score and the corresponding time in the region of $\xi_{n}^{e}$ are defined as $c_{n}^{e}$ and $t_{e,n}^{max}$, respectively. If $c_{n}^{s}$ or $c_{n}^{e}$ is higher than a threshold $\sigma\in\left[0,1\right]$, which is set manually for each specific dataset, we adjust the starting or ending point of $\xi_{n}$ with a weighting factor $\delta$ to control the contribution of $t_{s,n}^{max}$ and $t_{e,n}^{max}$ and yield the refined action instance $\xi_{n}^{\star}$. 
As such, the new  segment-level action instance set is refined to be  $\varphi_{p}^{\star}=\{\xi_{n}^{\star}\}_{n=1}^{M_{s}}$.

\textbf{Stage \rom{2}.} The middle probability sequence illustrates the probability of each frame whether it is inside one action proposal or not. We use the grouping scheme similar to TAG~\cite{TAG} to group the consecutive frames with high middle probability into regions as the candidate action instances. Such generated action instances are denoted by $\varphi_{tag}=\{\phi_{n}\}_{n=1}^{M_{t}}$ with $M_{t}$ indicating the total number of grouped action instances. 
We propose to make a further position adjustment by considering both  $\varphi_{tag}$ and  $\varphi_{p}^{\star}$.  Specifically, for each action instance $\xi_{n}^{\star}$ in $\varphi_{p}^{\star}$, its tIoU with all the action instances in $\varphi_{tag}$ are computed. If the maximum tIoU is higher than 0.8, the boundaries of $\xi_{n}^{\star}$ will be replaced by the corresponding action instance $\phi_{n}$ in $\varphi_{tag}$. Via such an operation, the substituted proposals are sensitive to boundaries and the overall boundary accuracy is improved accordingly.

\section{Experiments}

\subsection{Datasets} 

\textbf{THUMOS-14 \cite{thumos}.} It includes 1,010 videos and 1,574 videos with 20 action classes in the validation and test sets, respectively. There are 200 and 212 videos with temporal annotations of actions labeled in the validation and testing sets, respectively. We conduct the experiments on the same public split as~\cite{TAG,ctap}.

\textbf{ActivityNet-1.3 \cite{activitynet}.} The whole dataset consists of 19,994 videos
with 200 classes annotated, with 50\% for training, 25\% for validation, and the rest 25\% for testing. We train our model on the training set and perform evaluations on the
validation and testing sets, respectively.

\subsection{Temporal Proposal Generation}
In this section, we compare our proposed MGG against the existing state-of-the-art methods on both THUMOS-14 and ActivityNet-1.3 datasets.

For temporal action proposal, Average Recall (AR) computed with different tIoUs is usually adopted for performance evaluation. Following traditional practice, tIoU thresholds set from 0.5 to 0.95 with a step size of 0.05 are used on ActivityNet-1.3, while tIoU thresholds set from 0.5 to 1.0 with a step size of 0.05 are used on THUMOS-14. We also measure AR with different Average Numbers (ANs) of proposals, denoted as AR@AN. Moreover, the area under the AR-AN curve (AUC) is also used as one metric on ActivityNet-1.3, where AN ranges from 0 to 100.

Table~\ref{prop_thu} illustrates the performance comparisons on the testing set of THUMOS-14. Different feature representations will significantly affect the performances. As such, we adopt the two-stream~\cite{2stream} and C3D~\cite{c3d} features for fair comparisons. Taking the two-stream features as input, the AR@AN performances are consistently improved for AN ranging from 50 to 500, while BSN+NMS achieves a better performance with AN equal to 1000. While the C3D features are adopted, the AR@AN of MGG is higher than those of the other methods, with AN ranging from 50 to 1000. Such experiments clearly indicate the effectiveness of MGG in temporal proposal generation. 

\begin{table}
\scriptsize \centering \caption{\label{prop_thu}Performance comparisons with DAPs~\cite{dap}, SCNN-prop~\cite{shou_action}, SST~\cite{sst}, TURN~\cite{turn_tap}, BSN~\cite{bsn}, TAG~\cite{TAG}, and CTAP~\cite{ctap} on THUMOS-14 in terms of AR@AN.}
\label{tab:tab2}
\begin{tabular}{c|c|c|c|c|c|c}
\toprule Feature &Method &@50 & @100&@200 &@500& @1000\\\midrule
Flow & TURN& 21.86& 31.89& 43.02& 57.63&64.17\\
2-Stream & TAG& 18.55& 29.00& 39.61& -&-\\
2-stream & CTAP & 32.49 &42.61&51.97&-&-\\
2-Stream & BSN+NMS& 35.41& 43.55& 52.23& 61.35&\textbf{65.10}\\
2-Stream & \textbf{MGG}&\textbf{39.93} &\textbf{47.75} &\textbf{54.65} &\textbf{61.36} &64.06\\

\midrule
C3D&DAPs & 13.56& 23.83& 33.96&49.29 &57.64\\
C3D &SCNN-prop&17.22 & 26.17& 37.01& 51.57&58.20\\
C3D & SST& 19.90& 28.36& 37.90& 51.58&60.27\\
C3D & TURN& 19.63& 27.96& 38.34& 53.52&60.75\\
C3D & BSN+NMS& 27.19& 35.38& 43.61& 53.77&59.50\\
C3D & \textbf{MGG}&\textbf{29.11} &\textbf{36.31} & \textbf{44.32}&\textbf{54.95} &\textbf{60.98}\\

\bottomrule
\end{tabular}
\end{table}

\begin{figure}[]
\centering
\includegraphics[width=8cm]{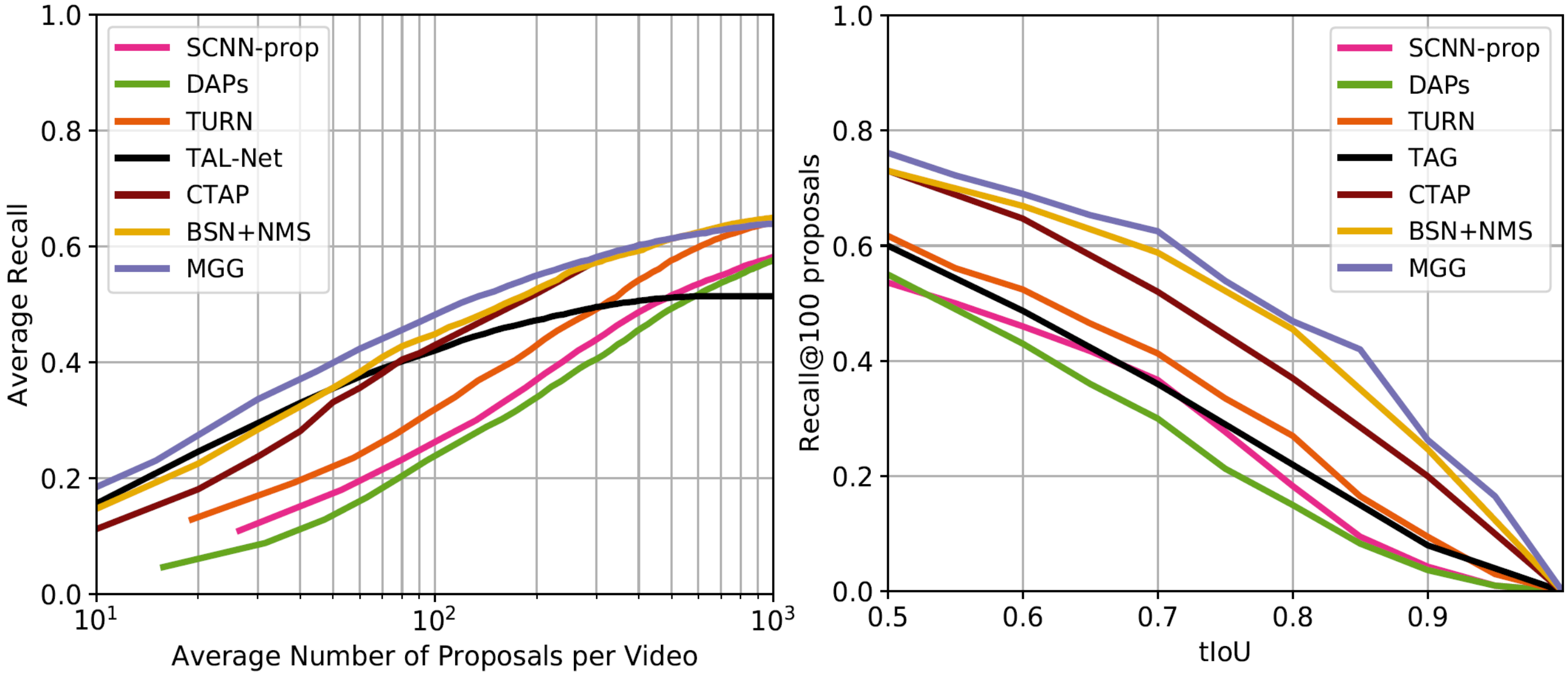}
\caption{AR-AN and recall@AN=100 curves of different temporal action proposal methods on the testing set of THUMOS-14.}
\label{thumos}
\end{figure}

Furthermore, Fig.~\ref{thumos} illustrates the AR-AN and recall@100-tIoU curves of different models on the testing split of THUMOS-14. It can be observed that our proposed MGG outperforms the other methods  in terms of AR-AN curves. Specifically, when AN equals 40, MGG significantly improves the performance from 33.02\% to 37.01\%. For recall@100-tIoU,
MGG gains a significantly higher recall when tIoU ranges from 0.5 to 1, indicating high accuracy of our proposal results.

\begin{table}
\scriptsize \centering \caption{\label{prop_act} Performance comparisons with TCN~\cite{tcn}, MSRA~\cite{yao}, Prop-SSAD~\cite{lin2017}, CTAP~\cite{ctap}, and BSN~\cite{bsn} on the validation and testing splits of ActivityNet-1.3.}
\label{tab:tab2}
\begin{tabular}{c|c|c|c|c|c|c}
\toprule Method &TCN &MSRA & Prop-SSAD&CTAP &BSN& \textbf{MGG}\\\midrule
AUC (val) &59.58 &63.12 & 64.40& 65.72& 66.17&\textbf{66.43}\\
AUC (test) & 61.56& 64.18& 64.80& -& 66.26&\textbf{66.47}\\
AR@100&- & -& 73.01& 73.17&74.16 &\textbf{74.54}\\
\bottomrule
\end{tabular}
\end{table}

Table~\ref{prop_act} illustrates the performance comparisons on the ActivityNet-1.3 dataset, where a two-stream ”Inflated 3D ConvNet” (I3D) model \cite{i3d} is used to extract features. Specifically, we compare our proposed MGG with the state-of-the-art methods, namely TCN~\cite{tcn}, MSRA~\cite{yao}, Prop-SSAD~\cite{lin2017}, CTAP~\cite{ctap}, and BSN~\cite{bsn}, in terms of AUC and AR@100. It can be observed that the proposed MGG outperforms the other methods on both the validation and testing sets. Specifically, MGG improves AR@100 on the validation set from 74.16 of the state-of-the-art method BSN to 74.54.

\begin{figure}
\centering
\includegraphics[width=8cm]{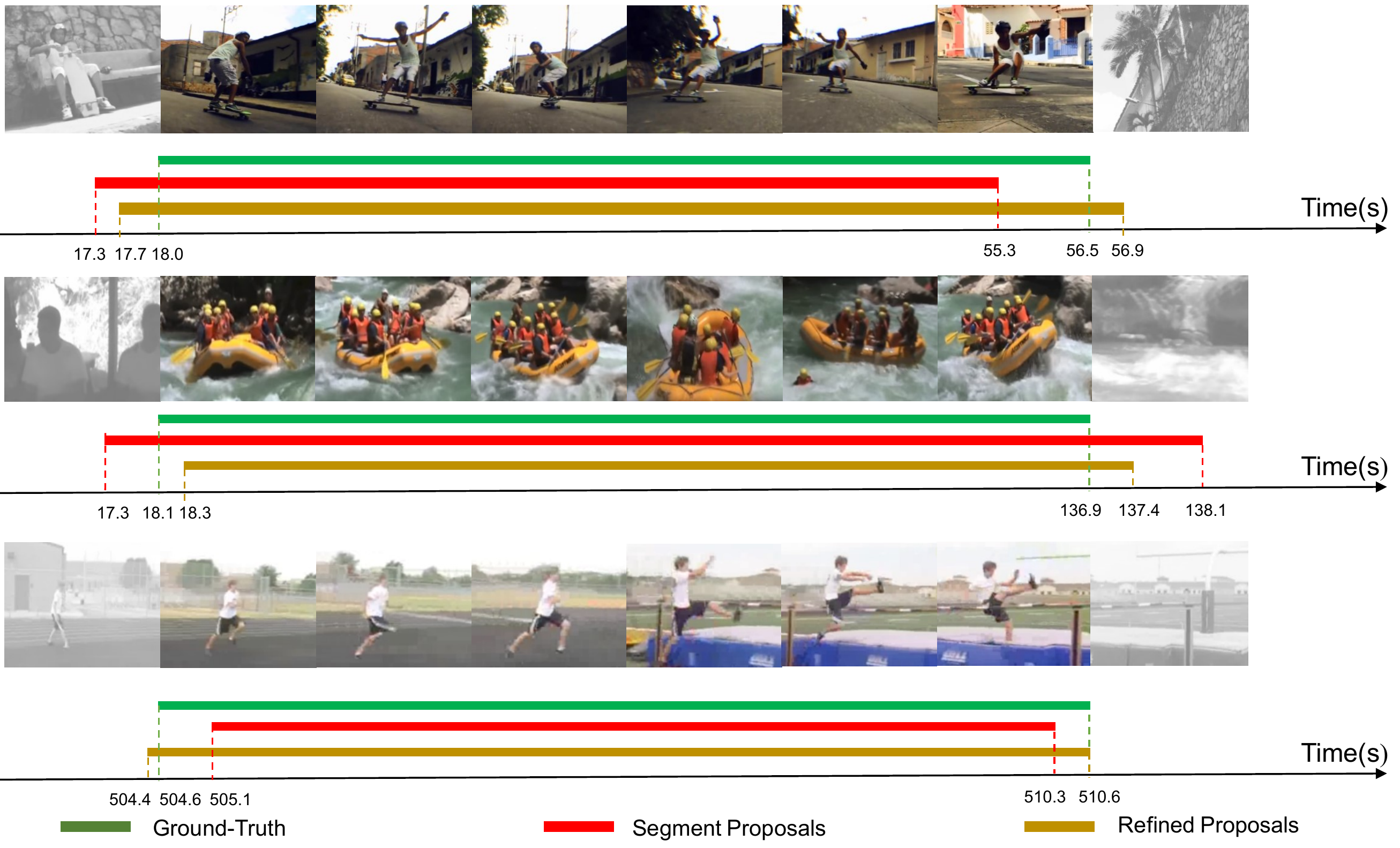}
\caption{\label{visual}Qualitative results of the proposals generated by MGG
on ActivityNet-1.3 (top and middle) and THUMOS-14 (bottom). It can be observed that the boundary information of the segment proposals 
generated by SPP is further adjusted using FAP, resulting in more precise proposals. }
\label{fig:architecure of network}
\end{figure}

Fig.~\ref{visual} illustrates some qualitative results of the generated proposals by MGG on ActivityNet-1.3 and THUMOS-14. Each is composed of  a sequence of frames sampled from a full video. By analyzing videos from both coarse and fine perspectives, MGG generates the refined proposals, with high overlapping with ground-truth proposals.

\subsection{Ablation Study}
In this subsection, the effect of each component in MGG is studied
in detail. We ablate the studies on the validation set of ActivityNet-1.3. Specifically, in order to verify the component effectiveness of MGG: position embedding, bilinear matching,  U-shape architecture in SPP, FAP, and SPP, we perform the ablation studies as follows:\\
   \textbf{MGG-P}: We discard the position information of the input video sequence and  directly feed the visual feature representations into MGG.\\
   \textbf{MGG-B}: We discard the bilinear matching model which exploits the interactions between the two temporal convolutions within BaseNet, and instead feed the output of the second convolutional layer to the following SPP and FAP.\\
  \textbf{MGG-U}: We discard the U-shape architecture which is proposed in SPP to increase semantic information of the lower layers. Correspondingly, only the expansive path of the feature pyramid is used.\\
   \textbf{MGG-F}: We only consider SPP to generate the final proposals, without considering FAP and the following TBA module.\\
   \textbf{MGG-S}: We only consider FAP to generate the final proposals, without considering SPP and the following TBA module.

As shown in Table \ref{ablation3_module}, our full model MGG outperforms all its variants, namely MGG-P, MGG-B, MGG-U, MGG-F, and MGG-S, which verifies the effectiveness of the components. In order to examine the detailed effectiveness of the U-shape architecture, we compare the recall rate of generated proposals in different lengths. As shown in Table~\ref{ablation_U}, the recall rate of short proposals drops dramatically, when the U-shape architecture is removed. The reason is that the U-shape architecture transfers higher semantic information to the lower layers, which can perceive global information of the video sequence, and is thus helpful for capturing proposals with short temporal extents.

\begin{table}
\centering \small \caption{\label{ablation3_module} 
Ablation studies on the validation set of ActivityNet-1.3 in terms of AUC and AR@AN.}
\begin{tabular}{l|c|c|c|c|c}
\toprule Method & AUC (val)&@30&@50&@80&@100\\\midrule
MGG-P &65.59&65.21&69.93&72.88&73.92\\
MGG-B & 65.88&65.56&70.41&73.19&73.89\\
MGG-U & 65.02&64.85&69.41&72.95&73.71\\
MGG-F & 64.31&63.76&67.91&71.04&72.24\\
MGG-S & 59.91    & 59.53    &  63.05   & 67.18    &  68.96    \\
\textbf{MGG} & \textbf{66.43}&\textbf{66.21}&\textbf{70.97}&\textbf{73.87}&\textbf{74.54}\\
\bottomrule
\end{tabular}
\end{table}

\begin{table}
\centering \scriptsize \caption{\label{ablation_U} 
Recall rates of MGG-U and MGG on generated proposals of different temporal extents on the validation set of ActivityNet-1.3, where AN and tIoU thresholds are set to 100 and 0.75, respectively.}
\begin{tabular}{l|c|c|c|c|c|c|c}
\toprule Method &0-5s&5-10s&10-15s&15-20s&25-30s&35-40s&40-45s\\\midrule
MGG-U & 0.15&0.63&0.73&0.80&0.91&0.93&0.94\\
MGG & \textbf{0.21}&\textbf{0.73}&\textbf{0.82}&\textbf{0.90}&0.93&0.93&0.92\\
\bottomrule
\end{tabular}
\end{table}

Moreover, it can be observed that MGG-F and MGG-S both perform inferiorly to our full MGG. The main reason is that SPP and FAP generate proposals at different granularities. Our proposed TBA can exploit their complementary behaviors and fuse them together to produce proposals with more precise boundary information. As introduced in Sec.~\ref{TBA}, TBA performs in two stages:\\
\textbf{Stage \rom{1}}: The starting and ending probability sequences generated by FAP are used to adjust boundaries of segment proposals from SPP.\\
\textbf{Stage \rom{2}}: The middle probability sequence is grouped into proposals with the method similar to \cite{TAG} and gives a final adjustment to boundaries of proposals from Stage \rom{1}.

Table~\ref{ablation_stage} illustrates the effectiveness of each stage in TBA. It can be observed that the two stages of TBA can both refine boundaries of segment proposals, thus consistently improving the performances, with AUC increasing from 64.31\% to 66.43\%.

\begin{table}
\centering \small \caption{\label{ablation_stage}
Performance comparisons of the two-stage TBA on the validation set of ActivityNet-1.3 in both end-to-end training and stagewise training manners. 
}\label{tab:tab2}
\begin{tabular}{l|ccc|ccc}
\hline
         & \multicolumn{3}{c|}{Stagewise}                                                    & \multicolumn{3}{c}{End-to-end} \\
\hline
MGG-F     & \checkmark & \checkmark & \checkmark & \checkmark & \checkmark & \checkmark \\
Stage \rom{1}   &            & \checkmark & \checkmark &          & \checkmark & \checkmark \\
Stage \rom{2}   &            &            &\checkmark &           &          & \checkmark \\
\hline
AUC(val) & 64.12      & 65.40      & 66.28    & 64.31      & 65.54     & 66.43\\
\hline
AR@100 &    72.05   & 73.41         &  74.19           &72.24        &    73.48     &74.54
\\
\hline
\end{tabular}
\end{table}

\textbf{Training: Stagewise v.s. End-to-end.} MGG is designed to jointly optimize SPP and FAP in an end-to-end fashion. It is also possible to train SPP and FAP separately, in which they do not work together. Such a training scheme is referred to as the stagewise training. Table~\ref{ablation_stage} illustrates the performance comparisons between end-to-end training and stagewise training. It can be observed that models trained in an end-to-end fashion can  outperform those learned with stagewise training under the same settings. It clearly demonstrates the importance of jointly optimizing SPP and FAP with BaseNet as a shared block to provide intermediate video representations.

\subsection{Action Detection}

In order to further examine the quality of generated proposals by MGG, we feed the detected proposals into the state-of-the-art action classifiers, including SCNN~\cite{shou_action} and UntrimmedNet~\cite{untrimmednet}. For fair comparisons, the same classifiers are also used for other proposal generation methods, including SST~\cite{sst}, TURN~\cite{turn_tap}, CTAP, and BSN.  We adopt the conventional mean Average Precision (mAP) metric, where Average
Precision (AP) reports the performance of each activity category. Specifically, mAP with
tIoU thresholds \{0.3, 0.4, 0.5, 0.6, 0.7\} is used on THUMOS-14.

Table~\ref{detect} illustrates the performance comparisons, which are evaluated on the testing set of THUMOS-14. With the same classifier, MGG achieves better performance than the other proposal generators, and outperforms the state-of-the-art proposal methods, namely CTAP~\cite{ctap} and BSN~\cite{bsn}, thus demonstrating the effectiveness of our proposed MGG.

\begin{table}
\scriptsize \centering \caption{\label{detect} Performance comparisons between MGG
and the other proposal generation methods in terms of video detection on the testing set of THUMOS-14, where mAP is reported with tIoU set from 0.3 to 0.7.
}\label{tab:tab2}
\begin{tabular}{c|c|ccccc}
\toprule Proposal Method &Classifier &0.7 & 0.6&0.5 &0.4&
0.3\\\midrule
SST \cite{sst}&SCNN-cls & -& -& 23.0& - &-\\
TURN \cite{turn_tap} &SCNN-cls &7.7 & 14.6& 25.6& 34.9&44.1\\
CTAP \cite{ctap} &SCNN-cls &- & -& 26.9& -&-\\
BSN \cite{bsn}&SCNN-cls &15.0 & 22.4& 29.4& 36.6&43.1\\
\textbf{MGG} & SCNN-cls& \textbf{15.8}& \textbf{23.6}& \textbf{29.9}& \textbf{37.8}&\textbf{44.9}\\
\hline
\hline
SST \cite{sst}& UNet& 4.7& 10.9& 20.0& 31.5&41.2\\
TURN \cite{turn_tap}& UNet& 6.3& 14.1& 24.5& 35.3&46.3\\
BSN \cite{bsn}&UNet &20.0 & 28.4& 36.9& 45.0&53.5\\
\textbf{MGG} & UNet& \textbf{21.3}& \textbf{29.5}& \textbf{37.4}& \textbf{46.8}&\textbf{53.9}\\
\bottomrule
\end{tabular}
\end{table}

\section{Conclusion}
In this paper, we proposed a novel architecture, namely MGG, for the temporal action proposal generation. MGG holds two branches: one is SPP perceiving the whole video in a coarse level and the other is FAP working in a finer level. SPP and FAP couple together and integrate into MGG, which can be trained in an end-to-end fashion. By analyzing whole videos from both coarse and fine perspectives,  MGG generates proposals with high recall and more precise boundary information. As such, MGG achieves better performance than the other state-of-the-art methods on the THUMOS-14 and ActivityNet-1.3 datasets. The superior performance of video detection relying on the generated proposals  further demonstrates the effectiveness of the proposed MGG.
\\
\\
\small{\textbf{Acknowledgements.} This work was supported in part by the Natural Science Foundation of Jiangsu under Grant BK20151102, in part by the State Key Laboratory for Novel Software Technology, Nanjing University under Grant KFKT2017B17, and in part by the Natural Science Foundation of China under Grant 61673108.}
{\small
\bibliographystyle{ieee}
\bibliography{egbib}
}

\clearpage
\appendix
\section{Supplementary Material}

This  supplementary material includes additional experiments that are not presented in the main paper and more qualitative results to demonstrate the performances of our proposed MGG.

\textbf{Recall Rates.}
Ground-truth proposals with short temporal spans are hard to capture, which mainly dues to that short proposals are of less semantic information. In the main paper, we illustrate the improvement of recall rates in short proposals with the U-shape architecture. Here, we demonstrate the short proposal recall rates of different methods including DAP \cite{dap}, TURN \cite{turn_tap}, CTAP \cite{ctap}, BSN \cite{bsn} and our proposed MGG on  the testing set of THUMOS-14. The temporal spans ranges from 1 frame to 60 frames, and the recall rates are computed with AN and tIoU set to 100 and 0.75, respectively. As shown in Table~\ref{recall}, the recall rates of MGG outperform the other competitor methods. One reason is that the temporal boundary adjustment (TBA) module is helpful for the proposal to be accurate in boundaries. Thus the generated proposals will have high overlap with ground-truths. Another reason is the U-shape architecture, which provides high-level semantic information for lower layers and helpful for the capture of proposals with short temporal durations.

\begin{figure*}[t]
\centering
\includegraphics[width=16cm]{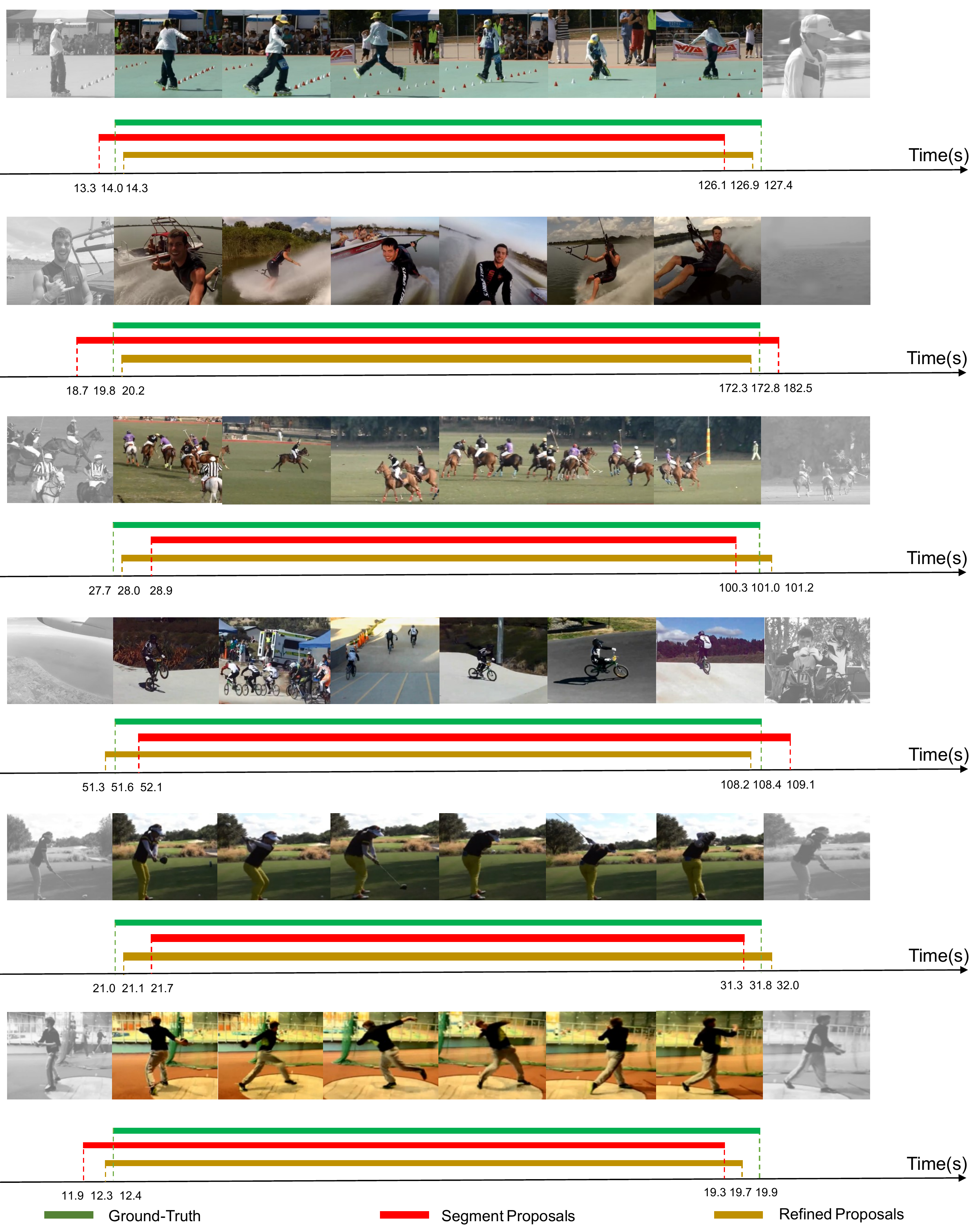}
\caption{Qualitative results of proposals generated by
MGG. First four rows represent temporal proposals on ActivityNet-1.3.
Last two rows represent temporal proposals on THUMOS-14.
After TBA  adopted to adjust proposal boundaries generated by segment proposal generator (SPG), the refined proposals will have high overlap with the ground-truth proposals.}
\label{fig:success}
\end{figure*}
\begin{figure*}[t]
\centering
\includegraphics[width=17cm]{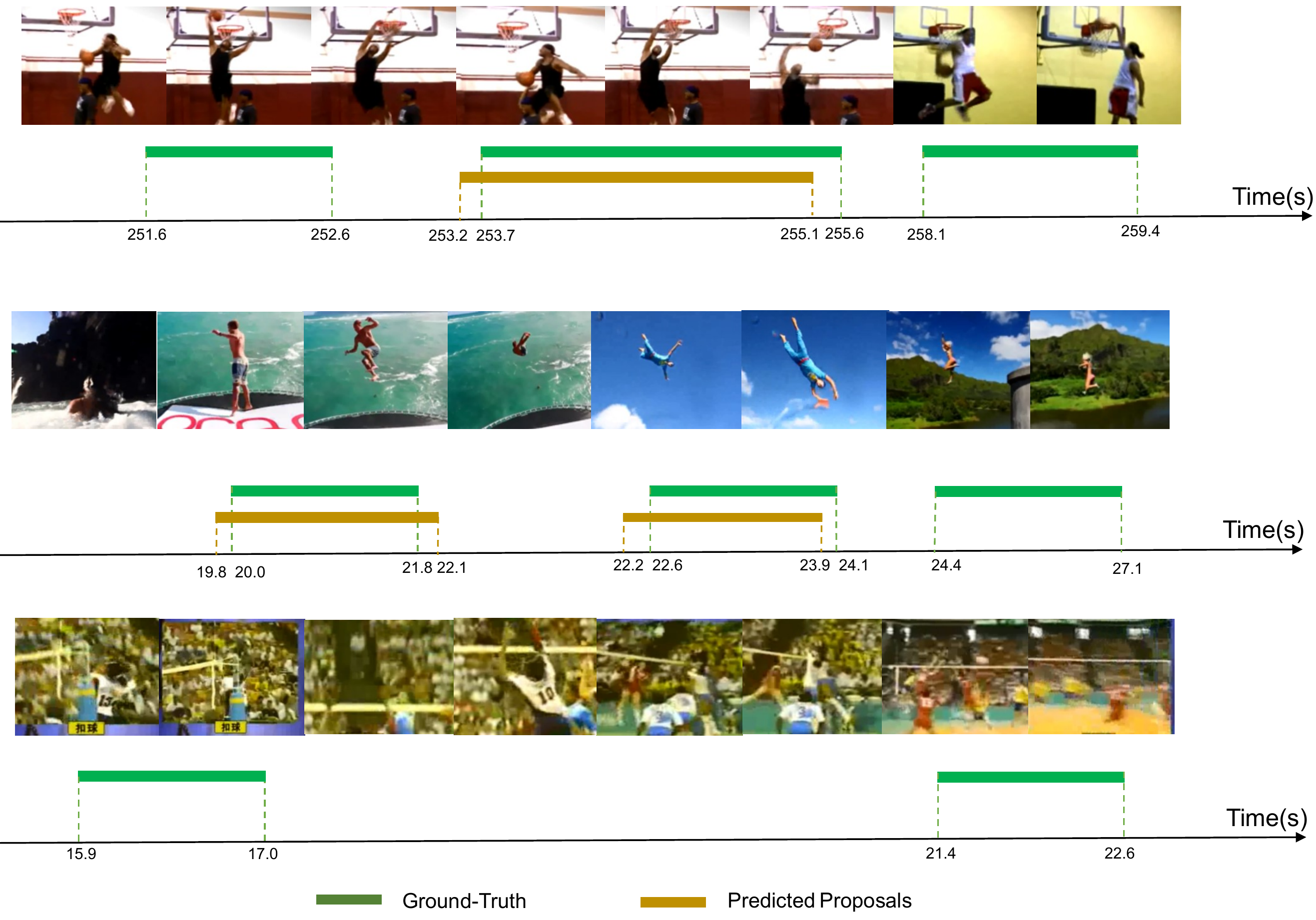}
\caption{Failure cases generated by
MGG on THUMOS-14.
For ground-truths with short temporal spans (first two rows), it is challenging for MGG to locate them. While quality of video frames is poorer (the last row), the performance will be reduced further.}
\label{fig:fail}
\end{figure*}

\textbf{Qualitative Results.}
More qualitative results are illustrated in Fig.~\ref{fig:success}. The first four rows are videos from the validation set of ActivityNet-1.3 \cite{activitynet} and the last two rows are from the testing set of THUMOS-14 \cite{thumos}. It can be observed that the refined proposals are of higher accuracy, which demonstrates the effectiveness of the proposed MGG.
Some failure cases are shown in Fig.~\ref{fig:fail}. For ground-truth proposals with short temporal durations, false negatives are produced. Moreover, if the videos are of low quality, it will be hard to capture the corresponding semantic meanings and thereby result in wrong proposals. 

\begin{table}[H]
\scriptsize
\caption{Recall rates of different methods on generated
proposals with short temporal extents on testing set of THUMOS-14, where AN and tIoU
thresholds are set to 100 and 0.75, respectively.}
\label{recall}
\begin{tabular}{c|cccccc}
\toprule
Method & 1-10 & 10-20 & 20-30 & 30-40 & 40-50 & 50-60 \\
\hline
DAP \cite{dap}                         & 0.000       & 0.000  & 0.025  & 0.047  & 0.097   & 0.106    \\
TURN \cite{turn_tap}                        & 0.000                      & 0.000  & 0.043  & 0.117  & 0.174   & 0.370    \\
CTAP \cite{ctap}                        & 0.000                      & 0.000  & 0.043  & 0.126  & 0.267   & 0.357    \\
BSN+NMS \cite{bsn}                        & 0.000                      & 0.048  & 0.168  & 0.237  & 0.339   & 0.400    \\
MGG                         & 0.000            & \textbf{0.081} & \textbf{0.183}& \textbf{0.296}& \textbf{0.364}&\textbf{0.431}    \\
\bottomrule
\end{tabular}
\end{table}

\end{document}